\documentclass[10pt,twocolumn,letterpaper]{article}

\usepackage{iccv}
\usepackage{times}
\usepackage{epsfig}
\usepackage{graphicx}
\usepackage{amsmath}
\usepackage{amssymb}

\usepackage{authblk}
\usepackage{multirow}
\usepackage[pagebackref=true,breaklinks=true,letterpaper=true,colorlinks,bookmarks=false]{hyperref}

\iccvfinalcopy % *** Uncomment this line for the final submission

\def\iccvPaperID{15} % *** Enter the ICCV Paper ID here

\ificcvfinal\fi
\begin{document}

%%%%%%%%% TITLE
\title{Convolutional Long Short-Term Memory Networks for Recognizing First Person Interactions}

%% omit the following lines up until the closing ``}''.
%% Additional authors and addresses can be added with ``\and'',
%% just like the second author.
%% To save space, use either the email address or home page, not both
%\and

\author[1,2]{Swathikiran Sudhakaran}
\author[2]{Oswald Lanz}
\affil[1]{University of Trento, Trento, Italy}
\affil[2]{Fondazione Bruno Kessler, Trento, Italy}
\affil[ ]{\tt\small {\{sudhakaran,lanz\}@fbk.eu}}

\maketitle
\thispagestyle{empty}

%%%%%%%%% ABSTRACT
\begin{abstract}
 In this paper, we present a novel deep learning based approach for addressing the problem of interaction recognition from a first person perspective. The proposed approach uses a pair of convolutional neural networks, whose parameters are shared, for extracting frame level features from successive frames of the video. The frame level features are then aggregated using a convolutional long short-term memory. The hidden state of the convolutional long short-term memory, after all the input video frames are processed, is used for classification in to the respective categories. The two branches of the convolutional neural network perform feature encoding on a short time interval whereas the convolutional long short term memory encodes the changes on a longer temporal duration. In our network the spatio-temporal structure of the input is preserved till the very final processing stage. Experimental results show that our method outperforms the state of the art on most recent first person interactions datasets that involve complex ego-motion. In particular, on UTKinect-FirstPerson it competes with methods that use depth image and skeletal joints information along with RGB images, while it surpasses all previous methods that use only RGB images by more than 20\% in recognition accuracy.
\end{abstract}

%%%%%%%%% BODY TEXT
\section{Introduction}
\label{sec:intro}

A new and vast array of research in the field of human activity recognition has been brought about with the recent technological advancements in wearable cameras. Majority of the human activity recognition techniques till now were concentrated on videos captured from a third person view. Developing techniques for the automatic analysis of ego-centric videos has immense application potential ranging from assistance during surgical procedure, law and order, assisting elderly citizens, video summarization, etc. But most of the existing research deal with activities carried out by the camera wearer, not interactions and reactions of others to the observer (person wearing the camera). The interaction recognition problem is different and difficult compared to the action recognition problem. This is because egocentric video captures a huge variety of objects, activities, and situations, and sometimes the person interacting with the observer can move out of the field of view as the person approaches the observer. Presence of ego-motion adds more complexity to the problem as it can interfere with the analysis of motion taking place in the scene. The applications of a method capable of automatically understanding interactions from egocentric view include surveillance, social interaction for robots, human machine interaction, etc.

Regardless of the recent advancements in deep learning techniques for addressing problems such as object recognition, caption generation, action recognition, etc., it has never been applied in the problem of first person interaction recognition. All the existing methods use hand-crafted features or processing stages for recognizing interactions in first person videos. End-to-end deep learning techniques require no prior information regarding the data and it is capable of achieving high degree of generalization compared to hand-crafted features based approaches. As a result, we propose to follow a deep learning approach for addressing the problem under study. To the best of our knowledge, this is the first time a deep learning based method is proposed for solving first person interaction recognition problem.

Our contributions can be summarized as follows:
\begin{itemize}
	\item We develop an end-to-end trainable deep neural network model for recognizing interactions from a first person perspective
	\item We propose a novel architecture that can utilize convolutional neural networks pre-trained on image datasets, for processing video data
	\item We experimentally evaluate the effectiveness of the proposed method on four publicly available first-person interaction recognition datasets
\end{itemize}

The remaining of the paper is organized as follows. A brief idea about the relevant state of the art techniques is provided in section \ref{sec:relWorks}. The proposed deep neural network architecture for addressing the concerned problem is discussed in section \ref{sec:propMeth}. Section \ref{expRes} discusses the experimental results obtained as part of the performance evaluation and the document is concluded in section \ref{concl}.

\begin{figure}[t]
	\centering
	\includegraphics[scale=0.18]{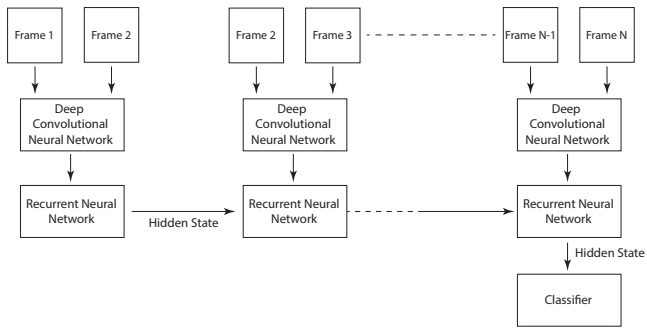}
	\caption{Basic block diagram showing the proposed method}
	\label{fig:gen_block}
\end{figure}

\section{Related Works}
\label{sec:relWorks}

An extensive amount of exploratory studies has been carried out in the area of first person activity or action recognition. They concentrate on the activity that the camera wearer is carrying out. These methods can be divided in to two major classes: activities involving object manipulation such as meal preparation \cite{fathi2011understanding,mccandless2013object,fathi2013modeling,matsuo2014attention,li2015delving,baraldi14gesture} and activities such as running, walking, etc. \cite{ryoo2015pooled,singh2016first,ma2016going,song2016egocentric,poleg2016compact}. The former relies on information about objects present in the scene for classifying the activity while the latter concentrates on the ego-motion and the salient motion in the scene.

More recent studies have focussed on the analysis of interaction recognition in the egocentric vision context. One of the pioneering work in this area is carried out by Ryoo \& Matthies in \cite{ryoo2013first}. They extracted optical flow based features and sparse spatio-temporal features from the videos and a bag of words model as the video feature representation. Narayan \emph{et al.} \cite{narayan2014action} use the TrajShape \cite{wang2013dense}, MBH \cite{dalal2006human} and HOF \cite{wang2013dense} as the motion features followed by bag of words approach or Fisher vector encoding \cite{perronnin2010improving} for generating the feature descriptor. Wray \emph{et al.} \cite{wray16sembed} propose graph-based semantic embedding for recognizing egocentric object interactions. Instead of focussing on the objects \cite{damen14modes}, Bambach \emph{et al.} \cite{bambach15hand} investigate the use of strong region proposals and CNN classifier to locate and distinguish hands involved in an interaction. 

Another line of research in this area was influenced by the development of Kinect device which is capable of capturing depth information from the scene together with the skeletal joints of the humans present in the scene. Methods exploiting this additional modality of data have been proposed by Gori \emph{et al.} \cite{gori2015robot}, Xia \emph{et al.} \cite{xia2015robot} and Gori \emph{et al.} \cite{gori2016multitype}. Gori \emph{et al.} \cite{gori2015robot} use the histogram of 3D joints, histogram of direction vectors and depth images along with the visual features. Xia \emph{et al.} propose to use the spatio-temporal features computed from the RGB and depth images \cite{xia2013spatio} together with the skeletal joints information as the feature descriptor. Gori \emph{et al.} \cite{gori2016multitype} propose a feature descriptor called relation history image which extracts information from skeletal joints and depth images.

Several deep learning based approaches have been developed by researchers for action recognition from third person view. Simonyan \& Zisserman \cite{simonyan2014two} use raw frames and optical flow images as input to two CNNs for extracting the feature descriptor. Donahue \emph{et al.} \cite{donahue2015long} and Srivastava \emph{et al.} \cite{srivastava2015unsupervised} use an architecture consisting of CNN followed by long short-term memory (LSTM) RNN for action recognition. A variant of the LSTM architecture, in which the fully-connected gates are replaced with convolutional gates (convLSTM), have been proposed by Shi \emph{et al.} \cite{xingjian2015convolutional} for addressing the problem of precipitation nowcasting prediction from radar images. The convLSTM is found to be functioning with improved performance compared to the fully-connected LSTM. The convLSTM model has been later used for predicting optical flow images \cite{PatrauceanHC16} and anomaly detection \cite{medel2016anomaly}. The results show that the convLSTM model suits applications involving spatio-temporal data such as videos. For this reason, we propose to use convLSTM as one of the important blocks in the proposed model for first person interaction recognition.

%------------------------------------------------------------------------
\section{Proposed Method}
\label{sec:propMeth}
Figure \ref{fig:gen_block} illustrates the basic idea behind the proposed approach. During each iteration two successive frames from the video will be applied to the model, i.e., frames 1 and 2 in the first time step, frames 2 and 3 in the second time step, etc. until all the video frames are inputted in to the network. The hidden state of the RNN in the final time step is then used as the feature representation of the video and is fed to the classifier.

Simonyan \& Zisserman \cite{simonyan2014two} have previously proposed to use optical flow images along with RGB images, as the inputs to CNN, for performing action recognition which has resulted in improved performance. The optical flow images will provide information regarding the motion changes in the frames which will in turn help the network in learning discriminating features that can distinguish one action from another. Inspired from this, we also evaluate the performance of the network when the difference between adjacent frames are applied as input. The difference of frames can be considered as an approximate version of optical flow images. Sun \emph{et al.} \cite{sun2015human} and Wang \emph{et al.} \cite{wang2016temporal} have also explored the possibility of using difference images as inputs to a deep neural network for action recognition. By applying frame difference as the input, the network is forced to learn the changes taking place in the video, i.e., the motion changes which define the actions and interactions occurring inside the video.

%The network functions in the following way when frame differences are applied as input. During each iteration, the difference between two pairs of successive frames are applied as input to the model. i.e., difference between frame 1 and frame 2, and the difference between frame 2 and frame 3 in the first time step; then the difference between frames 2 and 3 and the difference of frames 3 and 4 in the second time step and so on. Once all the frames are applied to the network in this manner, then the hidden state of the RNN in this final time step will be used as the feature representation of the entire video for further classification.

\begin{figure*}[t]
	\centering
	\includegraphics[width=0.95\textwidth]{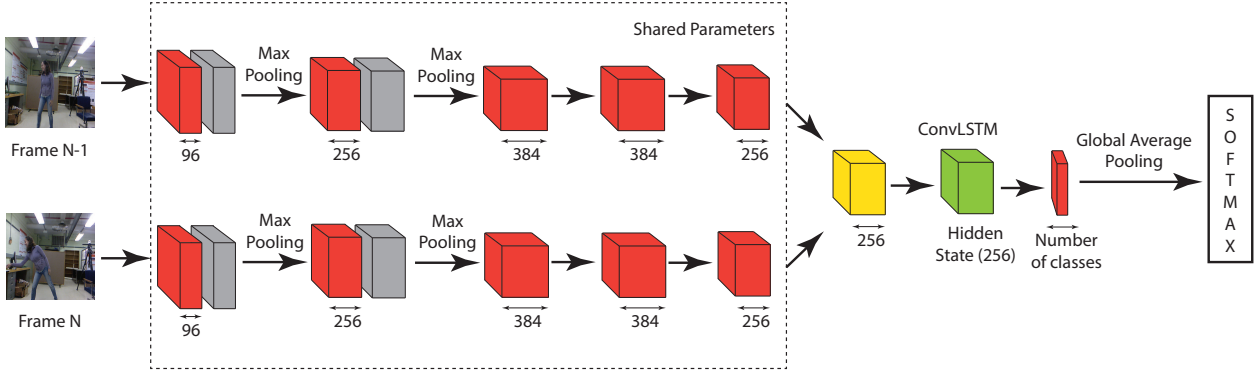}
	\caption{The architecture of the network. The convolutional layers are shown in red followed by normalization layer in gray. The 3D convolutional layer is shown in yellow and the convLSTM layer in green. We also experiment with a variant where, instead of raw frames we input difference-images obtained from pairs of successive frames.}
	\label{fig:block_dia}
\end{figure*}

\subsection{Convolutional Long Short-Term Memory}
\label{sec:convLSTM}
Convolutional long short-term memory (convLSTM) is an extension of the popular long short-term memory (LSTM) RNN \cite{hochreiter1997long}. In this the fully-connected gates of the LSTM module are replaced by convolutional gates thereby making it capable of encoding spatio-temporal features. The equations governing the functioning of convLSTM are:
\begin{eqnarray*}
i_t = \sigma(w_x^i* I_t + w_h^i*h_{t-1} + b^i)\\
f_t = \sigma(w_x^f* I_t + w_h^f*h_{t-1} + b^f)\\
\tilde{c}_t = tanh(w_x^{\tilde{c}}*I_t + w_h^{\tilde{c}}*h_{t-1} + b^{\tilde{c}})\\
c_t = \tilde{c}_t\odot i_t + c_{t-1}\odot f_t\\
o_t = \sigma(w_x^o * I_t + w_h^o*h_{t-1} + b^o)\\
h_t = o_t\odot tanh(c_t)
\end{eqnarray*}
where `*' and `$\odot$' represent the convolution operation and the Hadamard product. In this all the gate activations and the hidden states are 3D tensors as opposed to the case with standard fully-connected LSTM which are vectors.

\subsection{Network Architecture}

The architecture of the entire model used in the proposed approach is given in figure \ref{fig:block_dia}. The model can be divided in to the following parts:

\begin{enumerate}
	\item \textbf{Deep feature extractor}: A Siamese like architecture constituting of the AlexNet model \cite{krizhevsky2012imagenet} is used for extracting the frame level features. The fully-connected layers of the AlexNet model are removed. The two branches of the AlexNet model share their parameters and are pre-trained using the ImageNet dataset.
	\item \textbf{Spatio-temporal feature aggregator}: This consists of a 3D convolution layer followed by the convLSTM explained in section \ref{sec:convLSTM}. The output of the two branches of the previous module are concatenated together and is applied to the 3D convolution layer. The 3D convolutional layer uses a kernel of size $2\times3\times3$ and a stride of 1 in both the temporal and spatial dimensions. The output of the 3D convolution layer (256 feature planes) is then applied to the convLSTM layer for feature aggregation. The convLSTM layer uses 256 filters in all the gates with a filter size of $3\times3$ and stride 1. The hidden state, which contains the representation of the input video, consists of 256 feature planes.
	\item \textbf{Classifier}: Once all the input frames are applied to the first two stages of the network, the video representation, which is the hidden state of the convLSTM, is applied to the classifier. The classifier consists of a spatial convolution layer followed by global average pooling. The convolution layer generates an output with feature planes corresponding to the number of classes. It uses $3\times3$ kernels with stride 1. Global average pooling is performed on each of the feature map generated by the convolutional layer and is applied to the softmax layer. This is inspired from the network in network (NIN) model \cite{lin2013network} and the SqueezeNet model \cite{iandola2016squeezenet} proposed for image classification task.
\end{enumerate}

The two branches of the AlexNet extract relevant frame level information consisting of the objects present in the frame. This will help the network in understanding a frame level context of the video. By using two branches of the AlexNet together with the 3D convolution layer, information in a short temporal duration is encoded. An advantage of following this approach evolves from the fact that in this way existing powerful CNN architectures that are pre-trained on large scale image datasets can be used, along with the capability to take in to consideration the temporal context. The convLSTM layer maintains a memory of all the previous information it has seen which will result in the encoding of information in a longer extend of time. This is very relevant in the case of ego-centric videos. %This capability of the network to encode the short-term and long-term temporal information is relevant in the case of ego-centric videos. 
This is because, unlike in the case of third person views observing the scene from a distance, the objects or humans present in the video can move out of the field of view once they approach closer to the observer  for physical interaction, which necessitates the requirement for a memory.
Also, the convolutional gates present in the convLSTM helps the model to analyze the local spatial and temporal information together as compared to the fully-connected LSTM. This has a significant importance in the analysis of videos because the network should be capable of understanding the spatial and temporal changes occurring in the video. By removing the fully-connected layers of the AlexNet and using the convLSTM, the network achieves this capability. More specifically, we maintain spatial structure in the propagated memory tensor and generated video descriptor. This is in contrast e.g. to LRCN \cite{donahue2015long} and the many CNN-LSTM architectures that fully convolve into a vector before temporal aggregation with standard LSTM, propagating a memory vector. This late spatio-temporal aggregation is distinguishing and more effective in representing spatial structure in videos. Furthermore, prior to classification we maintain the spatial structure till the very final layer of our architecture by applying spatial convolution to obtain class-specific activations that are directly associated to local receptive fields in the input. By following the classifier used in the proposed approach, a significant reduction in the number of parameters is achieved thereby minimizing the propensity to overfit. On top of that, the global average pooling by itself can be considered as a regularization \cite{lin2013network}, thereby avoiding the usage of other regularization techniques.

\subsection{Training}
The implementation of the proposed method is done using the Torch library. We used RMSProp \cite{tieleman2012lecture} optimization algorithm for training the network and a fixed learning rate of $10^{-5}$ with a batch size of 12. The network is run for 10k iterations. The weights of the 3D convolution, convLSTM and the final spatial convolution layer are initialized using Xavier method \cite{glorot2010understanding}. The whole network is trained together to minimize the cross-entropy loss.

Since it is a generally accepted consensus that deep neural networks tend to overfit in the absence of large amount of training data, we used several data augmentation techniques for minimizing the possibility of overfitting. In the proposed approach, all the video frames are first rescaled to a dimension of $256\times340$. We follow the corner cropping and scale jittering techniques proposed by Wang \emph{et al.} \cite{wang2016temporal}. In this, during each training iteration, a portion of the frame (four corners or the centre) is cropped and is applied as input to the network. The dimension of the cropped region is selected randomly from the following set \{256, 224, 192, 168\}. The cropped region is then rescaled to $224\times224$. The selected regions are then horizontally flipped randomly during each training iteration. It is to be noted that the same augmentation techniques are applied to all the frames coming from a video during a particular training iteration. All the video frames in the training set are normalized to make their mean zero and variance unity.

%------------------------------------------------------------------------
\section{Experiments and Results}
\label{expRes}
The performance of the proposed approach is evaluated by testing it on four publicly available standard datasets. From all the videos, 20 frames equidistant in time are selected as the input to the network. The performance is evaluated in terms of recognition accuracy. During evaluation, the mean of the training frames are subtracted from the testing frames and is divided by the variance of the former. Image crops of dimension $224\times224$ are extracted from all four corners and the centre of each frame and are applied to the network. The horizontally flipped versions of the frames are also used during evaluation. Thus, for each video in the test set, 10 different samples (5 crops and their horizontal flips) are generated during inference. The outputs of all the crops and their flipped versions are then averaged together to find the class of the input video.

\begin{table*}[t!]
	\centering
	{\small\begin{tabular}{l| c| c| c| c}
		\hline
		\multirow{2}{*}{} & \multirow{2}{*}{JPLFPID} & \multirow{2}{*}{NUSFPID} & \multicolumn{2}{c}{UTKinect First Person Dataset}\\ \cline{4-5}
		& & & Humanoid & Non-humanoid\\
                \hline\hline
		Ryoo \& Matthies \cite{ryoo2013first}              &      89.6  &      -     &      57.1  &      58.48 \\
		Ryoo \cite{ryoo2011human}                          &      87.1  &      -     &      -     &      -     \\
		Abebe \emph{et al.} \cite{abebe2016robust}         &      86    &      -     &      -     &      -     \\
		Ozkan \emph{et al.} \cite{ozkan2017boosted}        &      87.4  &      -     &      -     &      -     \\
		Narayan \emph{et al.} \cite{narayan2014action}     & {\bf 96.7} &      61.8  &      61.9  &      57.6  \\
		Wang and Schmid \cite{wang2013action}              &      -     &      58.9  &      -     &      -     \\
		HOF \cite{ryoo2013first}                           &      -     &      -     &      45.92 &      -     \\
		Laptev \emph{et al.} \cite{laptev2008learning}     &      -     &      -     &      48.46 &      50.83 \\
                LRCN \cite{donahue2015long} (raw frames)           &      59.5  &      68.9  &      72.6  &      71.4  \\
                LRCN \cite{donahue2015long} (difference of frames) &      89.0  &      69.1  &      63.1  &      67.8  \\
                \hline
		{\bf Proposed Method} (raw frames)                 &      70.6  &      69.4  & {\bf 79.6} & {\bf 78.4} \\
		{\bf Proposed Method} (difference of frames)\hspace{.3in}       &      91.0  & {\bf 70.0} &      66.7  &      69.1  \\
                \hline
	\end{tabular}}\vspace{.1in}
	\caption{Comparison of the proposed method with existing techniques, on various datasets. Results are reported in terms of recognition accuracy in \%. \cite{narayan2014action} on UTKinect are our reproduced results following the paper description. LRCN \cite{donahue2015long} results are produced from the authors' code following same training/testing protocol used with our method. All other results are those available from the authors' papers.}
	\label{tab:comp1}
\end{table*}

\subsection{Datasets}
The following four datasets are used for validating the proposed first person interaction recognition approach.\\
\textbf{JPL First-Person Interaction Dataset (JPLFPID)} \cite{ryoo2013first} : The dataset consists of videos of humans interacting with a humanoid model with a camera attached on top of its head. The dataset consists of 7 different interactions which include friendly (hugging, shaking hands, etc.), hostile (punching and throwing) and neutral (pointing fingers) behaviors. There are 84 videos in the dataset and the evaluation method is done following existing approaches. During each evaluation instance, half of the videos are chosen as training and the remaining half as testing. Mean of the recognition accuracy obtained after each evaluation is reported as final accuracy.\\
\textbf{NUS First Person Interaction Dataset (NUSFPID)} \cite{narayan2014action} : This dataset, captured by placing the camera on the head of a human, consists of both actions (opening door, operating cell phone, etc.) and interactions (shaking hands, passing objects, etc.) with other humans from a first person view point. The dataset contains 152 videos. Random train-test split method is used for evaluation and the mean of the accuracies obtained after all the runs is used for reporting.\\
\textbf{UTKinect First Person Dataset-humanoid (UTKFPD-H)} \cite{xia2015robot} : In this dataset, a kinect sensor is mounted on a humanoid robot and the video of various humans interacting with it is captured. This dataset also contains friendly (hugging, shaking hands, etc.), hostile (punching, throwing objects, etc.) and neutral behavior (standing up, running, etc.) behaviors. There are 177 video samples from 9 different classes present in the dataset. The results reported are using the provided train-test split. For the experiments related to the proposed method, only the RGB videos are used and the depth information is not used.\\
\textbf{UTKinect First Person Dataset-non-humanoid (UTKFPD-NH)} \cite{xia2015robot} : This is similar to the above dataset with exception of the type of the robot used for capturing the video. A non-humanoid robot is used for capturing the videos in this dataset. The types of the interactions are also different from the previous dataset. There are 189 videos in the dataset and 9 different classes. Here also, the depth information available with the dataset is not used in the experiments.

All four datasets considered in this work is of significant difference from each other. For instance, the observer in each case is very much different so that the ego-motion that can occur to the video is different from one another. In the first dataset, the observer is stationary so that ego-motion can take place only if someone interacts with it physically. The NUSFPID and the UTKFPDH are more or less similar to each other but still the extend to which the interactions can affect their respective ego-motion is different, since the robot being more stable compared to a human. The type of interactions that are carried out in the datasets are also different. In this way, we did the experiments to evaluate the performance of the proposed method in identifying interactions with different levels of ego-motion and different variety of interactions.

\subsection{Results and Discussions}
The results obtained for each of the dataset is reported  and compared with the state of the art techniques in table \ref{tab:comp1}. As already mentioned, this is the first time a deep learning technique is proposed for recognizing first person interactions. Most of the existing deep learning based methods use gaze information or hand segmentations for identifying the objects being handled by the user which in turn is used for recognizing the action \cite{li2015delving,singh2016first}. Since first person interaction videos differ from such action recognition videos drastically, we are not considering these methods for comparing the performances. It is evident from the table that the proposed method is suitable for recognizing interactions from first person videos. Our method outperforms the state of the art methods except in one dataset (JPL first-person interaction dataset), where it came as the second best method. All the state of the art methods listed in table \ref{tab:comp1}, except LRCN, use hand-crafted features as opposed to the proposed approach which is based on deep learning. As mentioned in the previous section, the videos in the JPL first-person interaction dataset contain limited ego-motion as the camera is placed on a static object. Therefore, the ego-motion present in the videos is strongly correlated with the action that is occurring (for example, vertical motion when shaking hands, a sudden horizontal jerking motion in the punching action, etc.). Thus the ego-motion present in videos in the JPL dataset is part of the motion pattern that defines the action. In contrast to this, the rest of the datasets contain videos with strong ego-motion that may also arise occasionally and independently from the performed action. This may explain our findings that on the JPL dataset, we obtain improved performance when the difference of frames are applied as the input to the network since the difference operation will force the network to model the motion changes between frames (including the ego-motion that depends on the action class). On the other hand, the performance of the network on the UTKinect dataset is the highest when the raw frames are applied as input. The videos in the UTKinect dataset contains strong ego-motion. The presence of ego-motion that is uncorrelated with the action being performed will act as a sort of noise when the frame difference is taken as input: the network should be able to distinguish between the random ego-motion and the motion caused by the action, thereby making it more difficult for the network to isolate the salient motion patterns occurring in the video. The network performs in a comparable way on the NUSFPID dataset when either types of inputs are applied. Even though the NUSFPID dataset contains ego-motion, some of the classes include actions performed by the observer such as using cell phone, typing on the keyboard, writing, etc. For discriminating these types of actions that contain limited amount of motion, the network needs to learn the objects handled by the observer and hence the raw frames can be deemed as the most suited input modality. As opposed to this, interactions such as shaking hand, passing object, etc. contains motion and in this case, difference of frames performs the best. 

We have also compared the proposed method with another deep learning approach proposed for action recognition, LRCN \cite{donahue2015long}, in Table \ref{tab:comp1}. As mentioned previously, LRCN uses an AlexNet model for extracting frame level features followed by an LSTM for temporal aggregation. The results show that the proposed method is more suitable for first-person interaction recognition compared to the LRCN model. The primary reason is the usage of convLSTM in the proposed model as opposed to the fully-connected LSTM used in the LRCN. In LRCN, the input of the LSTM is the output of the penultimate fully-connected layer (fc6). The local spatial information in the video frame is lost by using the fully-connected layer, where as the proposed model preserves the spatial information by forwarding the convolutional features of the input. The convolutional gates of the convLSTM layer then performs information aggregation across both spatial and temporal domain. This is very important in the case of first-peron interaction recognition since it is required to memorize the changes occuring in both the spatial and temporal domains. Another feature of the proposed model that results in its improved performance is the utilization of the second AlexNet branch so that the model can process information from a broader temporal context. Since the parameters of both the branches are shared, there is no additional complexity added to the network and in addition to this, the same configuration can be used with any pre-trained CNN models. It should also be noted that, the proposed model achieves this improved performance with fewer number of parameters as compared to the LRCN model (21.8M ours vs 61.3M LRCN). This is achieved by eschewing from using fully-connected layers (both in the convLSTM and the classification layers). This will also enable the proposed model to be trained on smaller datasets without getting overfit.

Table \ref{tab:comp2} compares the performance of the proposed method on the UTKinect dataset with state of the art methods that use other modalities of information such as depth image and skeletal joints. It is clearly evident from the table that the proposed method performs comparably to these methods without using any other additional data. The proposed method is able to perform competitively to these methods by using only the RGB video as input.

\begin{table}[t]
	\centering
	{\small\begin{tabular}{l|c|c}
		\hline
		& UTKFPD-H & UTKFPD-NH\\
                \hline\hline 
		Xia \emph{et al.} \cite{xia2013spatio}      &   72.83  &   53.25 \\
		Oreifej \& Liu \cite{oreifej2013hon4d}      &   52.54  &   45.55 \\
		Xia \emph{et al.} \cite{xia2015robot}       &   85.6   &   83.7  \\
		Xia \emph{et al.} \cite{xia2012view}        &   -      &   70    \\
		Gori \emph{et al.} \cite{gori2016multitype} &   -      &   85.94 \\ \hline
		{\bf Proposed Method}\hspace{.3in}                       &   79.6   &   78.4  \\ \hline
	\end{tabular}}\vspace{.1in}
	\caption{Recognition accuracy obtained for the UTKinect first person dataset when the raw frames are applied as input to the proposed method. All other methods listed in the table uses either the depth image or skeletal joints information or both along with the RGB images for feature descriptor generation}
	\label{tab:comp2}
\end{table}

%------------------------------------------------------------------------
\section{Conclusions}
\label{concl}
The paper proposes a novel deep neural network architecture for recognizing interactions from first person videos. The proposed model extracts frame level features using a convolutional neural network followed by feature aggregation using a convolutional long short-term memory. Two different types of input modalities have been tested on the network. The first one uses adjacent frames as input and the second uses the difference between consecutive frames as input to the network. The proposed model is evaluated using four publicly available datasets. The network was able to outperform existing state of the art techniques. Raw frames is found as the suitable input to the network for videos with strong ego-motion, where as the difference of frames resulted in better performance for videos with less but action-correlated ego-motion. We also show that our proposed method performs competitively to the techniques that use depth image and skeletal joints information along with RGB images. 
%Experimental results show that the proposed method is able to outperform the state of the art techniques that use RGB images, by a significant margin, while performing competitively to the techniques that use depth image and skeletal joints information along with RGB images.
As a future work, we will explore the possibility of training the network jointly on both raw frames and difference of frames, thereby enabling the network to adapt by itself depending on the input videos, regardless of the presence or absence of strong ego-motion.

%------------------------------------------------------------------------
\section*{Acknowledgments}
We gratefully acknowledge the support of NVIDIA Corporation with the donation of GPU used for this research.

{\small
\bibliographystyle{ieee}
\bibliography{fpi}

\begin{thebibliography}{10}\itemsep=-1pt

\bibitem{abebe2016robust}
G.~Abebe, A.~Cavallaro, and X.~Parra.
\newblock Robust multi-dimensional motion features for first-person vision
  activity recognition.
\newblock {\em Computer Vision and Image Understanding}, 149:229--248, 2016.

\bibitem{bambach15hand}
S.~Bambach, S.~Lee, D.~J. Crandall, and C.~Yu.
\newblock Lending a hand: Detecting hands and recognizing activities in complex
  egocentric interactions.
\newblock In {\em ICCV}, 2015.

\bibitem{baraldi14gesture}
L.~Baraldi, F.~Paci, G.~Serra, L.~Benini, and R.~Cucchiara.
\newblock Gesture recognition in ego-centric videos using dense trajectories
  and hand segmentation.
\newblock In {\em CVPR Workshops}, 2014.

\bibitem{dalal2006human}
N.~Dalal, B.~Triggs, and C.~Schmid.
\newblock Human detection using oriented histograms of flow and appearance.
\newblock In {\em ECCV}, 2006.

\bibitem{damen14modes}
D.~Damen, T.~Leelasawassuk, O.~Haines, A.~Calway, and W.~Mayol-Cuevas.
\newblock You-do, i-learn: Discovering task relevant objects and their modes of
  interaction from multi-user egocentric video.
\newblock In {\em BMVC}, 2014.

\bibitem{donahue2015long}
J.~Donahue, L.~Anne~Hendricks, S.~Guadarrama, M.~Rohrbach, S.~Venugopalan,
  K.~Saenko, and T.~Darrell.
\newblock Long-term recurrent convolutional networks for visual recognition and
  description.
\newblock In {\em CVPR}, 2015.

\bibitem{fathi2011understanding}
A.~Fathi, A.~Farhadi, and J.~M. Rehg.
\newblock Understanding egocentric activities.
\newblock In {\em ICCV}, 2011.

\bibitem{fathi2013modeling}
A.~Fathi and J.~M. Rehg.
\newblock Modeling actions through state changes.
\newblock In {\em CVPR}, 2013.

\bibitem{glorot2010understanding}
X.~Glorot and Y.~Bengio.
\newblock Understanding the difficulty of training deep feedforward neural
  networks.
\newblock In {\em Aistats}, volume~9, 2010.

\bibitem{gori2016multitype}
I.~Gori, J.~Aggarwal, L.~Matthies, and M.~S. Ryoo.
\newblock Multitype activity recognition in robot-centric scenarios.
\newblock {\em IEEE Robotics and Automation Letters}, 1(1):593--600, 2016.

\bibitem{gori2015robot}
I.~Gori, J.~Sinapov, P.~Khante, P.~Stone, and J.~Aggarwal.
\newblock Robot-centric activity recognition ‘in the wild’.
\newblock In {\em International Conference on Social Robotics}, 2015.

\bibitem{hochreiter1997long}
S.~Hochreiter and J.~Schmidhuber.
\newblock Long short-term memory.
\newblock {\em Neural computation}, 9(8):1735--1780, 1997.

\bibitem{iandola2016squeezenet}
F.~N. Iandola, S.~Han, M.~W. Moskewicz, K.~Ashraf, W.~J. Dally, and K.~Keutzer.
\newblock Squeezenet: Alexnet-level accuracy with 50x fewer parameters and< 0.5
  mb model size.
\newblock {\em arXiv preprint arXiv:1602.07360}, 2016.

\bibitem{krizhevsky2012imagenet}
A.~Krizhevsky, I.~Sutskever, and G.~E. Hinton.
\newblock Imagenet classification with deep convolutional neural networks.
\newblock In {\em NIPS}, 2012.

\bibitem{laptev2008learning}
I.~Laptev, M.~Marszalek, C.~Schmid, and B.~Rozenfeld.
\newblock Learning realistic human actions from movies.
\newblock In {\em CVPR}, 2008.

\bibitem{li2015delving}
Y.~Li, Z.~Ye, and J.~M. Rehg.
\newblock Delving into egocentric actions.
\newblock In {\em CVPR}, 2015.

\bibitem{lin2013network}
M.~Lin, Q.~Chen, and S.~Yan.
\newblock Network in network.
\newblock {\em arXiv preprint arXiv:1312.4400}, 2013.

\bibitem{ma2016going}
M.~Ma, H.~Fan, and K.~M. Kitani.
\newblock Going deeper into first-person activity recognition.
\newblock In {\em CVPR}, 2016.

\bibitem{matsuo2014attention}
K.~Matsuo, K.~Yamada, S.~Ueno, and S.~Naito.
\newblock An attention-based activity recognition for egocentric video.
\newblock In {\em CVPR Workshops}, 2014.

\bibitem{mccandless2013object}
T.~McCandless and K.~Grauman.
\newblock Object-centric spatio-temporal pyramids for egocentric activity
  recognition.
\newblock In {\em BMVC}, 2013.

\bibitem{medel2016anomaly}
J.~R. Medel and A.~Savakis.
\newblock Anomaly detection in video using predictive convolutional long
  short-term memory networks.
\newblock {\em arXiv preprint arXiv:1612.00390}, 2016.

\bibitem{narayan2014action}
S.~Narayan, M.~S. Kankanhalli, and K.~R. Ramakrishnan.
\newblock Action and interaction recognition in first-person videos.
\newblock In {\em CVPR Workshops}, 2014.

\bibitem{oreifej2013hon4d}
O.~Oreifej and Z.~Liu.
\newblock Hon4d: Histogram of oriented 4d normals for activity recognition from
  depth sequences.
\newblock In {\em CVPR}, 2013.

\bibitem{ozkan2017boosted}
F.~Ozkan, M.~A. Arabaci, E.~Surer, and A.~Temizel.
\newblock Boosted multiple kernel learning for first-person activity
  recognition.
\newblock {\em arXiv preprint arXiv:1702.06799}, 2017.

\bibitem{PatrauceanHC16}
V.~P{\u a}tr{\u a}ucean, A.~Handa, and R.~Cipolla.
\newblock Spatio-temporal video autoencoder with differentiable memory.
\newblock In {\em ICLR Workshops}, 2016.

\bibitem{perronnin2010improving}
F.~Perronnin, J.~S{\'a}nchez, and T.~Mensink.
\newblock Improving the fisher kernel for large-scale image classification.
\newblock In {\em ECCV}, 2010.

\bibitem{poleg2016compact}
Y.~Poleg, A.~Ephrat, S.~Peleg, and C.~Arora.
\newblock Compact cnn for indexing egocentric videos.
\newblock In {\em WACV}, 2016.

\bibitem{ryoo2011human}
M.~S. Ryoo.
\newblock Human activity prediction: Early recognition of ongoing activities
  from streaming videos.
\newblock In {\em ICCV}, 2011.

\bibitem{ryoo2013first}
M.~S. Ryoo and L.~Matthies.
\newblock First-person activity recognition: What are they doing to me?
\newblock In {\em CVPR}, 2013.

\bibitem{ryoo2015pooled}
M.~S. Ryoo, B.~Rothrock, and L.~Matthies.
\newblock Pooled motion features for first-person videos.
\newblock In {\em CVPR}, 2015.

\bibitem{simonyan2014two}
K.~Simonyan and A.~Zisserman.
\newblock Two-stream convolutional networks for action recognition in videos.
\newblock In {\em NIPS}, 2014.

\bibitem{singh2016first}
S.~Singh, C.~Arora, and C.~Jawahar.
\newblock First person action recognition using deep learned descriptors.
\newblock In {\em CVPR}, 2016.

\bibitem{song2016egocentric}
S.~Song, N.-M. Cheung, V.~Chandrasekhar, B.~Mandal, and J.~Liri.
\newblock Egocentric activity recognition with multimodal fisher vector.
\newblock In {\em ICASSP}, 2016.

\bibitem{srivastava2015unsupervised}
N.~Srivastava, E.~Mansimov, and R.~Salakhutdinov.
\newblock Unsupervised learning of video representations using lstms.
\newblock In {\em ICML}, 2015.

\bibitem{sun2015human}
L.~Sun, K.~Jia, D.-Y. Yeung, and B.~E. Shi.
\newblock Human action recognition using factorized spatio-temporal
  convolutional networks.
\newblock In {\em ICCV}, 2015.

\bibitem{tieleman2012lecture}
T.~Tieleman and G.~Hinton.
\newblock Lecture 6.5-rmsprop: Divide the gradient by a running average of its
  recent magnitude.
\newblock {\em COURSERA: Neural networks for machine learning}, 4(2), 2012.

\bibitem{wang2013dense}
H.~Wang, A.~Kl{\"a}ser, C.~Schmid, and C.-L. Liu.
\newblock Dense trajectories and motion boundary descriptors for action
  recognition.
\newblock {\em International Journal of Computer Vision}, 103(1):60--79, 2013.

\bibitem{wang2013action}
H.~Wang and C.~Schmid.
\newblock Action recognition with improved trajectories.
\newblock In {\em ICCV}, 2013.

\bibitem{wang2016temporal}
L.~Wang, Y.~Xiong, Z.~Wang, Y.~Qiao, D.~Lin, X.~Tang, and L.~Van~Gool.
\newblock Temporal segment networks: towards good practices for deep action
  recognition.
\newblock In {\em ECCV}. Springer, 2016.

\bibitem{wray16sembed}
M.~Wray, D.~Moltisanti, W.~Mayol-Cuevas, and D.~Damen.
\newblock Sembed: Semantic embedding of egocentric action videos.
\newblock In {\em ECCV Workshops}, 2016.

\bibitem{xia2013spatio}
L.~Xia and J.~Aggarwal.
\newblock Spatio-temporal depth cuboid similarity feature for activity
  recognition using depth camera.
\newblock In {\em CVPR}, 2013.

\bibitem{xia2012view}
L.~Xia, C.-C. Chen, and J.~Aggarwal.
\newblock View invariant human action recognition using histograms of 3d
  joints.
\newblock In {\em CVPR Workshops}, 2012.

\bibitem{xia2015robot}
L.~Xia, I.~Gori, J.~K. Aggarwal, and M.~S. Ryoo.
\newblock Robot-centric activity recognition from first-person rgb-d videos.
\newblock In {\em WACV}, 2015.

\bibitem{xingjian2015convolutional}
S.~Xingjian, Z.~Chen, H.~Wang, D.-Y. Yeung, W.-K. Wong, and W.-c. Woo.
\newblock Convolutional lstm network: A machine learning approach for
  precipitation nowcasting.
\newblock In {\em NIPS}, 2015.

\end{thebibliography}
}

\end{document}